\documentclass[IEEEptsizenine]{IEEEtran}
\IEEEoverridecommandlockouts

\usepackage{cite}
\usepackage{url}
\usepackage{amsmath,amssymb,amsfonts}
\usepackage{algorithmic}
\usepackage{graphbox}
\usepackage{fontawesome}

\usepackage{multirow}
\usepackage{graphicx}
\usepackage{textcomp}
\usepackage{xcolor}
\def\BibTeX{{\rm B\kern-.05em{\sc i\kern-.025em b}\kern-.08em
    T\kern-.1667em\lower.7ex\hbox{E}\kern-.125emX}}

\usepackage{booktabs}

\newcommand{\new}[1]{{\color{black} {#1}}}

\usepackage{xspace}
\usepackage{hyperref}
\makeatletter
\DeclareRobustCommand\onedot{\futurelet\@let@token\@onedot}
\def\@onedot{\ifx\@let@token.\else.\null\fi\xspace}

\def\eg{\emph{e.g}\onedot} 
\def\ie{\emph{i.e}\onedot} 
\newcommand{\STAB}[1]{\begin{tabular}{@{}c@{}}#1\end{tabular}}

\begin{document}

\title{A Critical Assessment of Visual Sound Source Localization Models Including Negative Audio \\
\thanks{FPI scholarship PRE2022-101321, Maria de Maeztu CEX2021-001195-M/ AEI /10.13039/501100011033, MICINN/FEDER UE project ref. PID2021-
127643NB-I00, Fulbright Program and Ministerio de Universidades (Spain) funding for mobility stays of professors and researchers in foreign higher education and research centers.}
}

\author{
\IEEEauthorblockN{
Xavier Juanola$^{1}$,
Gloria Haro$^{1}$, 
Magdalena Fuentes$^{2}$
}\\
\IEEEauthorblockA{
    $^1$ Universitat Pompeu Fabra, Barcelona, Spain,\\
    $^2$ MARL-IDM, New York University, New York, USA\\
    }
}

\maketitle

\begin{abstract}
The task of Visual Sound Source Localization (VSSL) involves identifying the location of sound sources in visual scenes, integrating audio-visual data for enhanced scene understanding. Despite advancements in state-of-the-art (SOTA) models, we observe three critical flaws: i) The evaluation of the models is mainly focused in sounds produced by objects that are visible in the image,  ii) The evaluation often assumes a prior knowledge of the size of the sounding object, and iii) No universal threshold for localization in real-world scenarios is established, as previous approaches only consider positive examples without accounting for both positive and negative cases.
In this paper, we introduce {extended} test sets and {new} metrics designed to complete the current standard evaluation of VSSL models by testing them in scenarios where none of the objects in the image corresponds to the audio input, \ie  a \textit{negative} audio. We consider three types of \textit{negative} audio: \textit{silence}, \textit{noise} and \textit{offscreen}. Our analysis reveals that numerous SOTA models fail to appropriately adjust their predictions based on audio input, suggesting that these models may not be leveraging audio information as intended.
Additionally, we provide a comprehensive analysis of the range of maximum values in the estimated audio-visual similarity maps, in both \textit{positive} and \textit{negative} audio cases, and show that most of the models are not discriminative enough, making them unfit to choose a universal threshold appropriate to perform sound localization without any a priori information of the sounding object, that is, object size and visibility.
\end{abstract}


\section{Introduction}
\label{sec:intro}

Visual Sound Source Localization (VSSL) aims at identifying the location of  the sounding objects in a visual scene. It is a key aspect in scene understanding with applications in scene monitoring \cite{surveillance}, robot navigation \cite{robotics_application} and post-processing for enhancing content accessibility \cite{gao_grauman_coseparation}, among others.

Pioneering works \cite{fisher2000, HersheyMovellan1999, pixels_that_sound} addressed VSSL by localizing the sound sources in a given image using classical machine learning techniques. Since then, many learning-based models have been proposed. 
Most of the works in the literature are assessed in scenarios with a single sounding object that is visible in the input image, \eg \cite{arandjelovic2017look, arandjelovic2018objects, LVS, owens2018audio, SSLTIE, EZ-VSL, SLAVC, senocak2018learning, senocak2024aligning, FNAC, RCGrad}. This is not true in practical scenarios where it is common to encounter silent objects, {sounds emitted by objects out of the image frame (i.e. offscreen sounds)} and different sources sounding at the same time (\ie a mixture of sounds). 

Although substantial progress has been made in VSSL in the last years regarding audio mixtures \cite{hu2022mix, kim_mixtures_2024,  mahmud2024mixture, mo2023mixture, qian2020mixture}, only a few attempted to work with silent objects \cite{DSOL, liu2022visual} or offscreen sounds \cite{liu2022visual}.
Most importantly, the majority of existing works evaluate the models with just \emph{positive} audio examples, that is, with sounds (in a mixture or not) produced by objects that are \textit{visible} in the image. Except for \cite{SLAVC}, no analysis is performed using \emph{negative} audio, \ie audio sources that are not visible. By evaluating with only positive cases, current approaches fail to represent real-world audio-visual scenarios and risk drawing misleading conclusions about how models interpret the audio input. While \cite{SLAVC} considers silence and offscreen sounds, it does not provide metrics to analyse the \textit{negative} sounds separately from the \textit{positive} ones, making it difficult to understand the model's behaviour,
and the proportion of audio-visual pairs with a \textit{negative} audio  is small compared to those with \textit{positive} audio. 
Another shortcoming of current methods is that when evaluating, they either assume that the object is visible \cite{senocak2018learning} and/or that their size is known \cite{senocak2024aligning}, which is typically not the case in practice.

We propose a more comprehensive evaluation of VSSL models, that does not assume any prior information about the sounding objects and that includes not only \textit{positive} audio examples but also different types of \textit{negative} audio, namely: \textit{silence}, \textit{noise} and \textit{offscreen} sounds. Consequently, we propose new metrics that assess the performance both in the \textit{negative} cases and globally in \textit{positive} and \textit{negative} cases. By examining the overlap of model prediction maps for both \textit{positive} and \textit{negative} cases, we identify which models are more discriminative and can be better adapted to work in real-world applications via an absolute threshold.
Our contributions are the following: i) Two {extended} test sets for evaluating VSSL models that includes both \textit{positive} and  \textit{negative} audio examples; 
ii) New metrics to evaluate the performance of VSSL models on \textit{negative} audio alone and to assess the global performance both on \textit{positive} and \textit{negative} audio;
iii) A comprehensive analysis of VSSL models on their discriminatory power between sounding and non sounding objects, for which we identify model-specific universal thresholds to perform VSSL without any prior information of the sounding object related to its size and its visibility.
Our code is open source and available in \url{https://github.com/xavijuanola/vssl_eval}.

\section{Method}
\label{sec:method}

\noindent {\bf Extended test datasets.} To assess the performance of VSSL models both with \textit{positive} and \textit{negative} audio  we extend two existing testing datasets: the VGG-SS test set \cite{LVS} and the recently proposed synthetic IS3 dataset \cite{senocak2024aligning}. 
{For that,} we include three types of  \textit{negative} audio: \textit{silence}, \textit{noise} and \textit{offscreen} sounds. We pair every image in the test set with its \textit{positive} audio {(the original one)}  and the three types of \textit{negative} audio {that we add}. 
For negative cases, the models' localization map (\ie a binarized map derived from likelihood values assigned to image pixels indicating the presence of a sounding object) should be empty (\ie low values).

VGG-SS features 221 categories and IS3 130 categories which are a subset from the VGG-SS ones. However, there are many categories highly related semantically: \eg there are 6 categories related to dog (\eg `dog barking', `dog bow-wow' or `dog howling'), 3 related to train (`train wheels squealing', `train whistling' and `train horning'), 2 related to woman (`female singing' and `female speech, woman speaking'), and so on. These categories are impossible to distinguish from each other within the visual modality when we just consider a single image. 
Motivated by this, we decided to define a set of broad categories for both datasets in order to select a true offscreen audio for a given category that also belongs to a completely different semantic category.
The 8 broad semantic  categories that we propose are the following: \textit{music/musical instruments},  \textit{human voice}, \textit{vehicles},  \textit{devices/machines}, \textit{animals}, \textit{weapons},  \textit{nature ambient sounds} and \textit{others/miscellaneous}. 
The categories are distinct enough (\eg \emph{animals} vs. \emph{instruments} instead of \emph{cats} vs. \emph{dogs}) and thus \textit{offscreen} sounds are easily distinguishable from onscreen sounds so that a reasonable VSSL model should be able to discriminate them.

In the case of \textit{noise}, we generate a sequence of random values with normal distribution with zero mean  and standard deviation 1. {These values are then clipped to the range [-1, 1].} For \textit{silence}, we pair the images with an empty audio.
Our extended VGG-SS constitutes a total of 5,537 images, 4.61 hours of \textit{positive} audio and 13.84 hours of \textit{negative audio}.
For the extended IS3 we have 3,242 distinct images and 6,484 audio files, corresponding to 5.40 hours of \textit{positive} audio (one audio for each of the two visible sources in each image) and 16.21 hours of \textit{negative} audio. 
For a fair comparison, we ensure that all models are tested against the same \textit{noise} and \textit{offscreen} audio samples. Given that the randomness in the  assignments of offscreen sounds and noise generation can affect our experiments, we repeat them 10 times with different seeds and report the average metrics.

\medskip 
\noindent {\bf Metrics for Positive and Negative cases.} Since we propose to evaluate the models with both \textit{positive} and \textit{negative} audio inputs, we need metrics to assess the performance in each case.

For \textit{positive} audio we use the standard metrics in the literature, proposed in \cite{senocak2018learning}: consensus Intersection over Union (cIoU) and Area under the Curve (AUC). The localization accuracy is measured by cIoU, which is the intersection over union between a ground-truth map computed as a consensus among different annotators and the estimated localization map. The localization map is usually obtained by binarizing the audio-visual similarity map -- computed as the per pixel cosine similarity of visual and audio embeddings -- by a certain threshold \cite{LVS, EZ-VSL, FNAC, SLAVC, SSLTIE, senocak2024aligning}.  Instead, RCGrad \cite{RCGrad}, uses gradCAM \cite{gradCAM} to generate the localization maps. As noted in \cite{senocak2024aligning}, the common threshold that activates the top 50\% pixels is not adapted to different sizes of sounding objects. Thus, they propose to use an adaptive threshold that activates the top $M$ pixels, where $M$ is the area of the ground truth object. The cIoU computed with this adaptive threshold is denoted as cIoU Adap. However, this adaptive threshold is assuming two types of prior information about the sounding object: i) that the object is visible and ii) that its size is known. Finally, AUC measures the integral of the success ratio (proportion of samples with cIoU $\geq \tau$)  as a function of the threshold $\tau$ varying from 0 to 1. AUC Adap.\ is the analogous to AUC but using cIoU Adap.\ instead.

To assess the performance of models on \textit{negative} audio samples and ensure a comprehensive understanding of their behaviour, we introduce the following new metrics: 

\subsubsection{pIA}
\label{sec:pia}
It is designed to evaluate the performance of models on \textit{negative} audio samples as it measures the relative percentage of image area that has been activated in the model's localization map. We denote the metric as \textit{percentage of Image Area}, defined as
$
    \text{pIA} = \sum_x m(x) / (H \cdot W),
$
where $m \in \{0, 1\}^{H \times W}$ is 
the localization map. We have that $0 \leq \text{pIA} \leq 1$, and the lower the value the better.

\subsubsection{AUC$_{\text{N}}$}
\label{sec:aoc}

This metric is analogous to AUC for the case of \textit{negative} audio, and it uses the pIA. The success ratio in case of \textit{negative} audio is the proportion of samples with  $\text{pIA} \leq \tau$. Then, AUC$_{\text{N}}$ measures the integral of the success ratio in \textit{negative} audio samples varying the threshold $\tau$ from 0 to 1. In this case the higher the metric, the better.

\subsubsection{F$_{\text{LOC}}$ and F$_{\text{AUC}}$} 
\label{sec:harmonic_mean}

For the purpose of globally assessing the models' performance in a single metric, we propose to use the harmonic mean of the  metrics for both positive and negative cases.    
We define metrics as follows:
$ \text{F}_{\text{LOC}} = 2 \cdot \text{cIoU} \cdot \left(1 - \overline{\text{pIA}}\right)/\left(\text{cIoU} + 1 - \overline{\text{pIA}}\right)$, and 
$ \text{F}_{\text{AUC}} =
2 \cdot \text{AUC} \cdot \overline{\text{AUC}}_\text{N} / \left(\text{AUC} + \overline{\text{AUC}}_\text{N} \right)$, 
where $\overline{\text{pIA}}$ and $\overline{\text{AUC}}_\text{N}$ are  the mean of the pIA and AUC$_\text{N}$ for the three \textit{negative} cases.

\subsubsection{$IoU$ between two localization maps}
\label{sec:iou_maps}

This metric computes the Intersection over Union (IoU) of two different localization maps, and helps to measure how similar are the localizations of two different audio inputs. 
We report four different values: i) \textit{positive} vs.~\textit{silence}, ii) \textit{positive} vs.~\textit{noise}, iii) \textit{positive} vs.~\textit{offscreen}, and iv) \textit{negative} vs.~\textit{negative} (average of the pairwise IoU between negatives). Ideally, the localization map of a \textit{negative} sound is an empty map, so there should not be any overlap between the localization of the \textit{positive} audio and the localization of a \textit{negative} one. Thus, the IoU in cases i), ii) and iii) should be zero. In the case of iv), the ideal situation would be to have two empty localization maps. To avoid an undefined value when two localization maps are empty, we set the IoU to one (\ie we consider them as equal).

\begin{table*}[ht]
\caption{Quantitative results on the VGG-SS and IS3 \new{extended} test sets.  
Best results in {\bf bold}, second best \underline{underlined}.}
\label{tab:all_metrics}
\centering
\resizebox{\textwidth}{!}{
\begin{tabular}{@{}clccccccccccccccc@{}}\toprule
& & \multicolumn{4}{c}{{\bf Positive audio input}} & \phantom{abc} & \multicolumn{6}{c}{{\bf Negative audio input}} & & & \multicolumn{2}{c}{{\bf Global metric}} \\
& &  & &  &  & & \multicolumn{2}{c}{{\bf Silence}} & \multicolumn{2}{c}{{\bf Noise}} & \multicolumn{2}{c}{{\bf Offscreen sound}} & &  &  &\\
\cmidrule{3-6} \cmidrule{8-9} \cmidrule{10-11} \cmidrule{12-13} \cmidrule{16-17}
{\bf Test set} & {\bf Model} & {\bf cIoU \new{U. th.} $\uparrow$} & {\bf cIoU Adap. $\uparrow$} & {\bf AUC \new{U. th.} $\uparrow$} & {\bf AUC Adap. $\uparrow$} & & {\bf pIA $\downarrow$} & {\bf AUC$_{\text{N}}$ $\uparrow$} & {\bf pIA $\downarrow$}& {\bf AUC$_{\text{N}}$ $\uparrow$} & {\bf pIA $\downarrow$}& {\bf AUC$_{\text{N}}$ $\uparrow$} & & & {\bf F$_{\text{LOC}}$ $\uparrow$}  & {\bf F$_{\text{AUC}}$ $\uparrow$} \\
\midrule

\multirow{7}{*}{\STAB{\rotatebox[origin=c]{90}{\textbf{\new{Extended} VGG-SS}}}} 
 & RCGrad \cite{RCGrad} 
    & \small 11.71 
    & \small 37.05 
    & \small 12.07 
    & \small 37.26 
    & 
    & \small 4.42 
    & \small 95.80 
    & \small 4.43 
    & \small 95.80 
    & \small 4.41 
    & \small 95.81 
    & 
    & 
    & \small 20.86 
    & \small 20.86 \\

 & LVS \cite{LVS} 
    & \small 2.26
    & \small 38.59 
    & \small 4.45 
    & \small 40.38 
    & 
    & \small 1.51 
    & \small 98.39 
    & \small \textbf{0.07} 
    & \small \textbf{99.91} 
    & \small \textbf{0.54} 
    & \small \textbf{99.39} 
    & 
    & 
    & \small 4.43 
    & \small 8.52\\

 & EZ-VSL \cite{EZ-VSL} 
     & \small 12.79 
     & \small 44.74 
     & \small 14.28 
     & \small 43.08 
     & 
     & \small \textbf{0.34} 
     & \small \textbf{99.63} 
     & \small 0.34 
     & \small 99.63 
     & \small 1.86
     & \small 98.05
     & 
     & 
     & \small 22.66 
     & \small 24.96\\

 & FNAC \cite{FNAC} 
    & \small 12.74 
    & \small 48.03 
    & \small 14.26 
    & \small 44.35 
    & 
    & \small 3.81 
    & \small 96.19 
    & \small 3.81 
    & \small 96.19 
    & \small 2.70 
    & \small 97.27 
    & 
    & 
    & \small 22.50 
    & \small 24.84\\
 
 & SLAVC \cite{SLAVC} 
    & \small \underline{18.28} 
    & \small 49.64 
    & \small \underline{19.54} 
    & \small 45.90
    & 
    & \small 0.84
    & \small 99.16
    & \small 0.84
    & \small 99.16
    & \small 4.73
    & \small 95.23
    & 
    & 
    & \small \underline{30.81} 
    & \small \underline{32.57}\\

 & SSL-TIE \cite{SSLTIE} 
    & \small \textbf{18.67} 
    & \small \underline{49.86}
    & \small \textbf{19.57} 
    & \small \textbf{47.05} 
    & 
    & \small \underline{0.52} 
    & \small \underline{99.42} 
    & \small 0.45 
    & \small 99.52 
    & \small 1.98 
    & \small 97.93 
    & 
    & 
    & \small \textbf{31.41} 
    & \small \textbf{32.68}\\

 & SSL-Align \cite{senocak2024aligning} 
    & \small 16.65 
    & \small \textbf{53.10} 
    & \small 17.79 
    & \small \underline{46.62}
    & 
    & \small 1.05 
    & \small 98.89 
    & \small \underline{0.29}
    & \small \underline{99.67} 
    & \small \underline{1.79}
    & \small \underline{98.09}
    & 
    & 
    & \small 28.50 
    & \small 30.16\\

\midrule 

\multirow{7}{*}{\STAB{\rotatebox[origin=c]{90}{\textbf{\new{Extended} IS3}}}} 
 & RCGrad \cite{RCGrad} 
    & \small 0.10 
    & \small 1.07 
    & \small 2.54 
    & \small 3.57 
    & 
    & \small 4.49 
    & \small 95.81 
    & \small 4.49 
    & \small 95.79 
    & \small 4.47 
    & \small 95.83 
    & 
    & 
    & \small 0.20 
    & \small 0.20\\

 & LVS \cite{LVS} 
    & \small 1.95 
    & \small 39.38 
    & \small 4.19 
    & \small 41.07 
    & 
    & \small 1.02 
    & \small 98.85 
    & \small \textbf{0.03}
    & \small \textbf{99.96} 
    & \small \textbf{0.20} 
    & \small \textbf{99.77}
    & 
    & 
    & \small 3.82 
    & \small 8.04\\

 & EZ-VSL \cite{EZ-VSL} 
    & \small 3.47 
    & \small 42.08 
    & \small 5.55 
    & \small 42.66 
    & 
    & \small \underline{0.30} 
    & \small \underline{99.67} 
    & \small 0.30 
    & \small 99.67 
    & \small \underline{0.70} 
    & \small \underline{99.22} 
    & 
    & 
    & \small 6.72 
    & \small 10.51\\

 & FNAC \cite{FNAC} 
    & \small 10.03 
    & \small 49.01 
    & \small 11.76 
    & \small 45.71 
    & 
    & \small 3.27 
    & \small 96.69 
    & \small 3.27 
    & \small 96.69 
    & \small 2.00 
    & \small 97.97 
    & 
    & 
    & \small 18.18 
    & \small 20.97\\
 
 & SLAVC \cite{SLAVC} 
    & \small 10.42 
    & \small 45.05 
    & \small 12.16 
    & \small 42.96 
    & 
    & \small 0.68 
    & \small 99.29 
    & \small 0.68
    & \small 99.29 
    & \small 4.03
    & \small 95.95 
    & 
    & 
    & \small 18.83 
    & \small 21.64\\

 & SSL-TIE \cite{SSLTIE} 
    & \small \underline{17.39} 
    & \small \underline{51.53} 
    & \small \underline{17.39} 
    & \small \underline{47.49} 
    & 
    & \small \textbf{0.28} 
    & \small \textbf{99.66} 
    & \small \underline{0.23} 
    & \small \underline{99.72} 
    & \small 2.05 
    & \small 97.85 
    & 
    & 
    & \small \underline{29.58} 
    & \small \underline{31.15}\\

 & SSL-Align \cite{senocak2024aligning} 
    & \small \textbf{22.54} 
    & \small \textbf{63.10} 
    & \small \textbf{23.51} 
    & \small \textbf{52.40} 
    & 
    & \small 1.86 
    & \small 97.98 
    & \small 0.45 
    & \small 99.49 
    & \small 0.82
    & \small 99.11
    & 
    & 
    & \small \textbf{36.72} 
    & \small \textbf{37.99}\\
\bottomrule
\end{tabular}%
}
\end{table*}

\medskip 
\noindent {\bf Universal Threshold for localization in the wild.} As mentioned above, the common practice of assessing the localization accuracy in \textit{positive} audio involves a binarizing threshold that is adapted to each image and implies some prior information. However, in order to evaluate the models in an ``in-the-wild'' scenario -- with no assumptions on the sounding objects and better reflecting the practical use of VSSL models -- we need a universal threshold, common for all images.

\begin{figure}[htb]
   \centering
   
\includegraphics[width=\columnwidth]{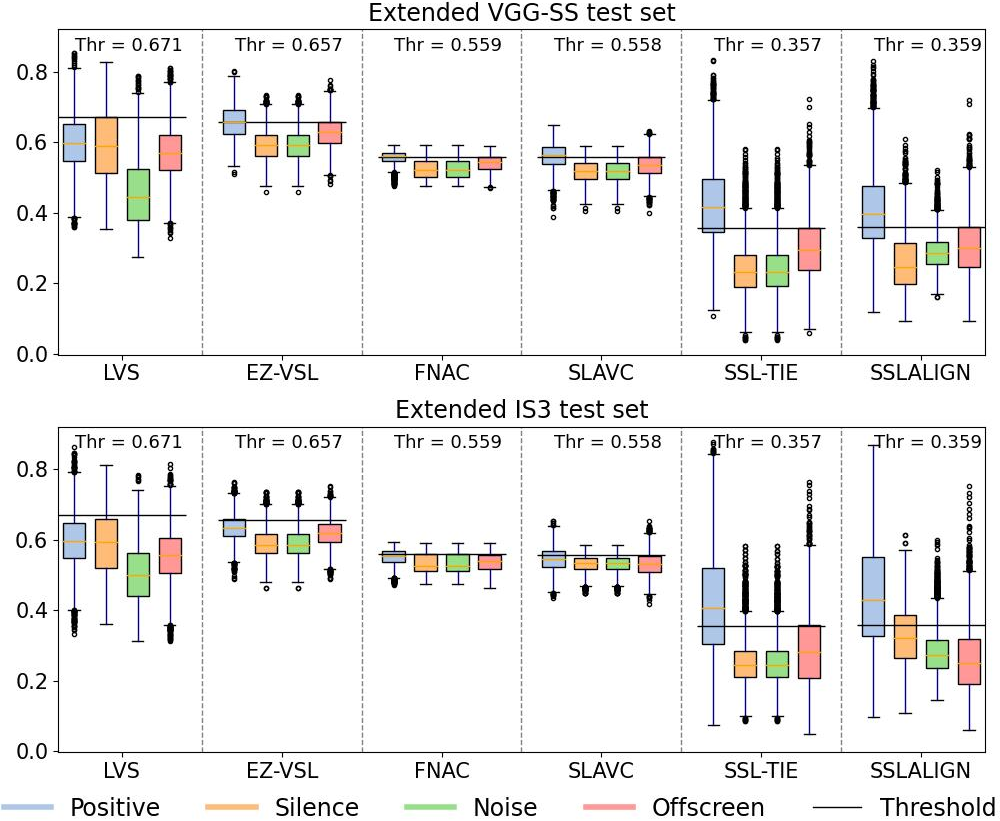}
\caption{Distribution of the max. values of the audio-visual cosine similarities per different models and types of audio (depicted in different colors) across the two different \new{extended} test sets: VGGSS (top) and IS3 (bottom).
The universal threshold value (Thr) for every model is shown with a solid line.}
\label{fig:boxplot}
\end{figure}

With the aim of automatically choosing an adequate universal threshold for each model, we analyse in Fig. \ref{fig:boxplot} the distribution of the maximum value in the audio-visual similarity map (\ie before binarization), both for the samples in VGG-SS and IS3 test sets. Four types of audio samples for each image in the test set are used: \textit{positive}, \textit{silence}, \textit{noise} and \textit{offscreen}. We focus only on the maximum value because, in a positive case, it should exceed the threshold, while in a negative case, it should not (since no pixel in the image ``sounds"). 
Fig. \ref{fig:boxplot} shows a boxplot of the distribution of the maximum value, where the colored area inside the box represents the  Inter Quartile Range (IQR), \ie the second and third quartiles, where 50\% of the values are concentrated. This figure illustrates the discriminatory power of each model  in distinguishing between \textit{positive} and \textit{negative} audio. We analyze the following VSSL models: RCGrad \cite{RCGrad}, LVS \cite{LVS}, EZ-VSL \cite{EZ-VSL}, SLAVC \cite{SLAVC}, SSL-TIE \cite{SSLTIE} and SSL-Align \cite{senocak2024aligning}. 

As it can be observed in Fig. \ref{fig:boxplot}, most models present a strong overlap in the range of maximum similarities for the cases of \textit{positive}, \textit{silence} and \textit{offscreen} audio, thus making it difficult to choose a universal threshold in which the negative audios will not be activated in the localization map. 
SSL-TIE and SSL-Align are the ones that present a higher discriminatory power. SSL-TIE is the most discriminative model in VGG-SS. In IS3, SSL-TIE is good at discriminating \textit{silence} and \textit{noise} but less good in case of \textit{offscreen}. On the other hand, SSL-Align is good  at discriminating \textit{noise} and \textit{offscreen} in IS3 but less good at \textit{silence}. In light of the results in Fig. \ref{fig:boxplot}, we decide to set the universal localization threshold for each model as the highest 75th percentile, or 3rd quartile, of the maximum similarity values of the three types of \textit{negative} audio.
This value is computed in the \new{extended} test set of VGG-SS and is shown   in Fig. \ref{fig:boxplot} (both in numbers and as a horizontal line). We use this threshold to binarize the audio-visual similarity maps and thus obtain the localization map used in all metrics. This threshold is used in the \new{extended} test sets of both VGG-SS and IS3; in this latter case with the aim of assessing if it generalizes.
The threshold is computed differently for RCGrad model, as it uses gradCAM, with output values normalized in the interval $[0, 1]$. In this case, we use a threshold of $0.75$.

\section{Experiments}
\label{sec:experiments}

\begin{figure}[htb]
   \centering
\begin{tabular}{cccccc}
& \scriptsize \hspace{-0.35cm} \raisebox{-0.17cm}{\includegraphics[height=0.5cm]{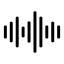}} Man & \scriptsize \hspace{-0.5cm} \raisebox{-0.17cm}{\includegraphics[height=0.5cm]{Figures/equalizer.png}} Ukulele & \scriptsize \hspace{-0.5cm} \raisebox{-0.17cm}{\includegraphics[height=0.5cm]{Figures/equalizer.png}} Silence & \scriptsize \hspace{-0.6cm} \raisebox{-0.17cm}{\includegraphics[height=0.5cm]{Figures/equalizer.png}} Noise & \hspace{-0.52cm} \raisebox{-0.17cm}{\includegraphics[height=0.5cm]{Figures/equalizer.png}} \scriptsize Offscreen \\
\vspace{0.05cm}
\rotatebox[origin=c]{90}{\scriptsize RCGrad } & 
\hspace{-0.45cm} \includegraphics[align=c, width=1.54cm]{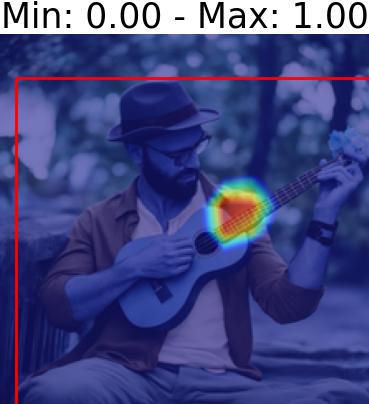}&
\hspace{-0.45cm} \includegraphics[align=c, width=1.54cm]{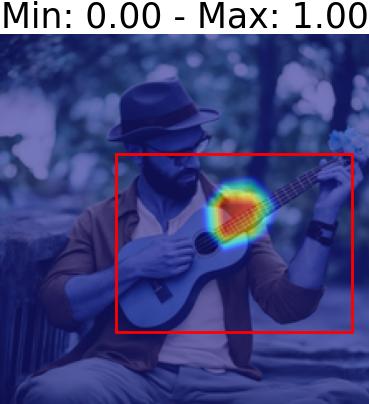}&
\hspace{-0.45cm} \includegraphics[align=c, width=1.54cm]{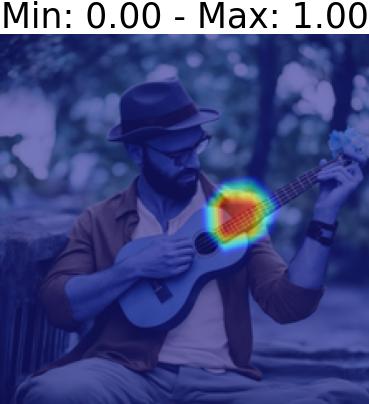}&
\hspace{-0.45cm} \includegraphics[align=c, width=1.54cm]{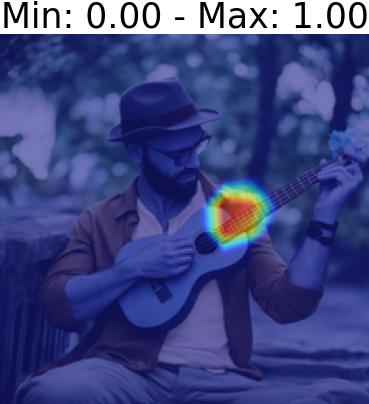}&
\hspace{-0.45cm} \includegraphics[align=c, width=1.54cm]{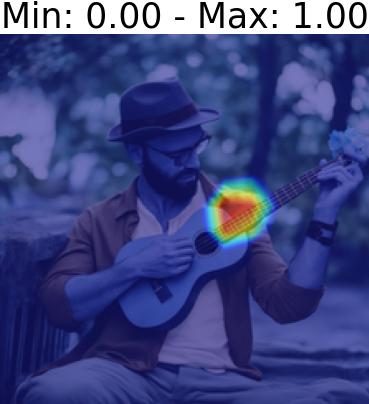}\\
\vspace{0.05cm}
\rotatebox[origin=c]{90}{\scriptsize LVS } & 
\hspace{-0.45cm} \includegraphics[align=c, width=1.54cm]{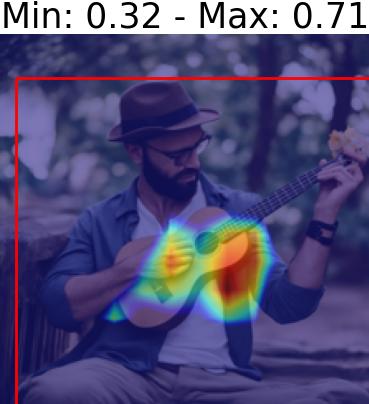}&
\hspace{-0.45cm} \includegraphics[align=c, width=1.54cm]{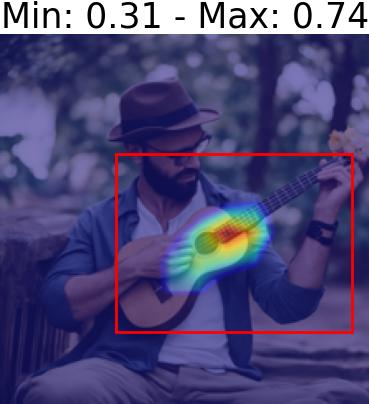}&
\hspace{-0.45cm} \includegraphics[align=c, width=1.54cm]{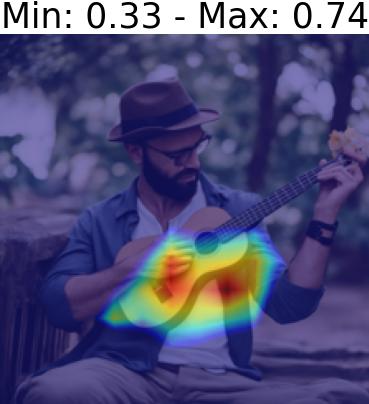}&
\hspace{-0.45cm} \includegraphics[align=c, width=1.54cm]{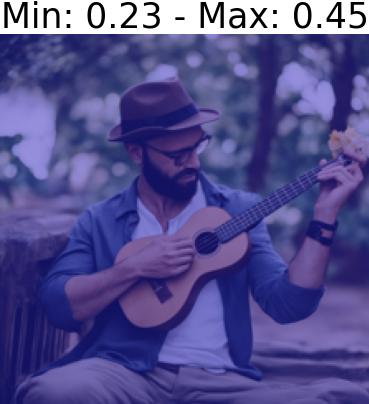}&
\hspace{-0.45cm} \includegraphics[align=c, width=1.54cm]{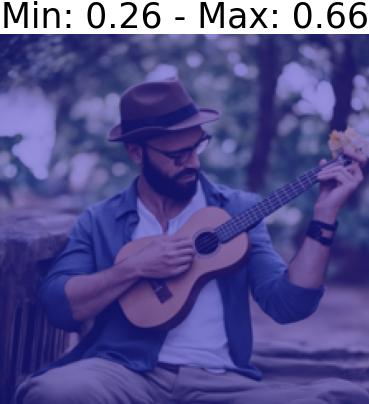}\\
\vspace{0.05cm}
\rotatebox[origin=c]{90}{\scriptsize EZ-VSL } & 
\hspace{-0.45cm} \includegraphics[align=c, width=1.54cm]{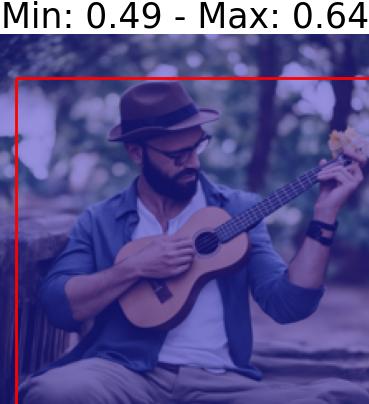}&
\hspace{-0.45cm} \includegraphics[align=c, width=1.54cm]{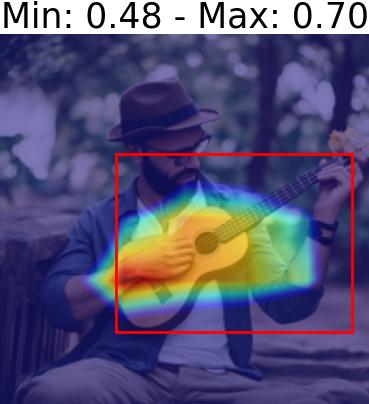}&
\hspace{-0.45cm} \includegraphics[align=c, width=1.54cm]{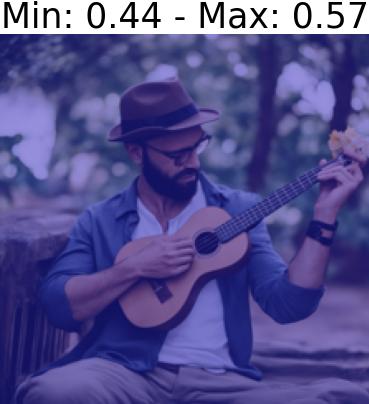}&
\hspace{-0.45cm} \includegraphics[align=c, width=1.54cm]{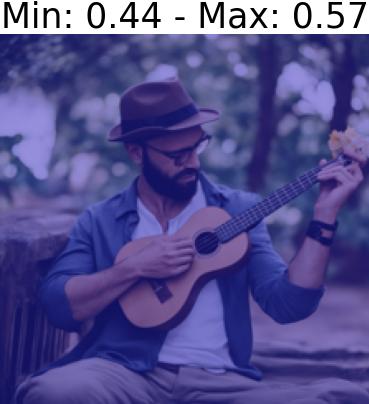}&
\hspace{-0.45cm} \includegraphics[align=c, width=1.54cm]{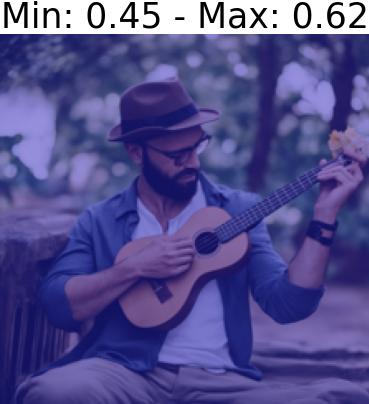}\\
\vspace{0.05cm}
\rotatebox[origin=c]{90}{\scriptsize FNAC } & 
\hspace{-0.45cm} \includegraphics[align=c, width=1.54cm]{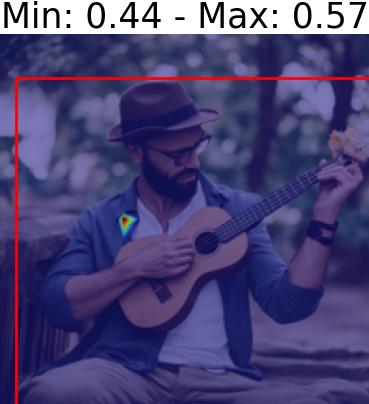}&
\hspace{-0.45cm} \includegraphics[align=c, width=1.54cm]{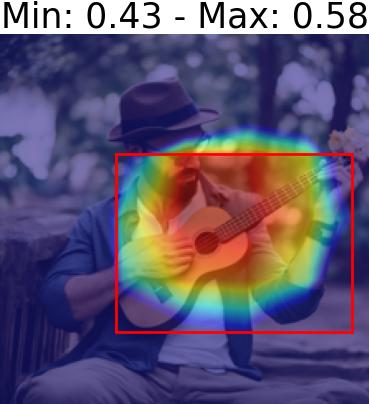}&
\hspace{-0.45cm} \includegraphics[align=c, width=1.54cm]{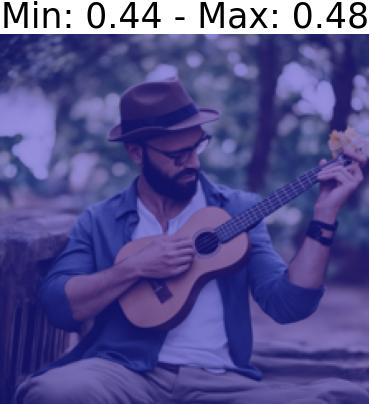}&
\hspace{-0.45cm} \includegraphics[align=c, width=1.54cm]{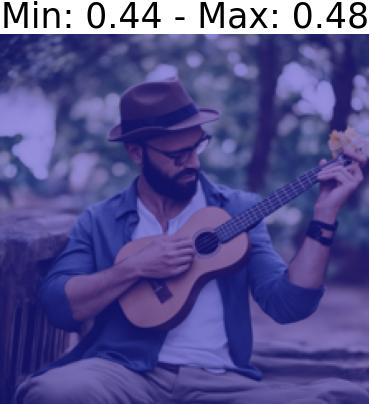}&
\hspace{-0.45cm} \includegraphics[align=c, width=1.54cm]{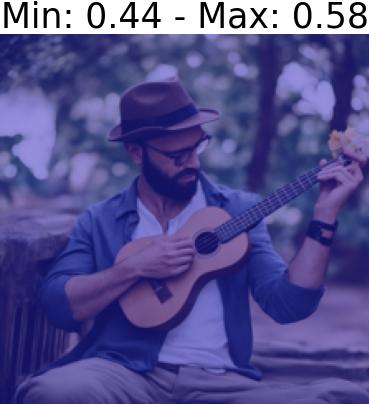}\\
\vspace{0.05cm}
\rotatebox[origin=c]{90}{\scriptsize SLAVC } & 
\hspace{-0.45cm} \includegraphics[align=c, width=1.54cm]{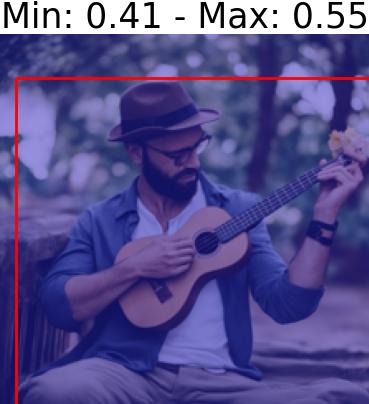}&
\hspace{-0.45cm} \includegraphics[align=c, width=1.54cm]{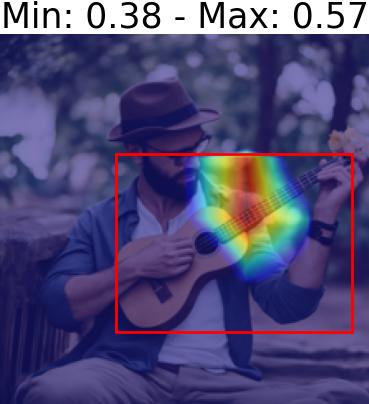}&
\hspace{-0.45cm} \includegraphics[align=c, width=1.54cm]{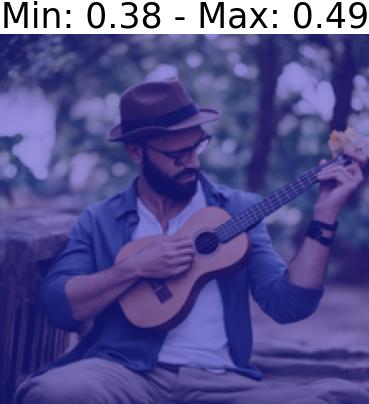}&
\hspace{-0.45cm} \includegraphics[align=c, width=1.54cm]{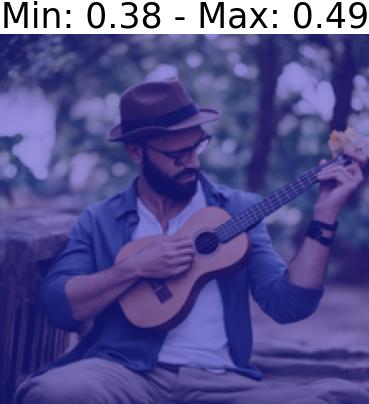}&
\hspace{-0.45cm} \includegraphics[align=c, width=1.54cm]{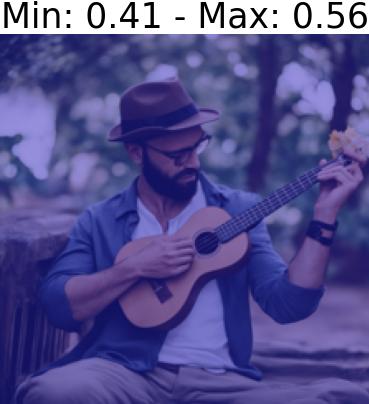}\\
\vspace{0.05cm}
\rotatebox[origin=c]{90}{\scriptsize SSL-TIE } & 
\hspace{-0.45cm} \includegraphics[align=c, width=1.54cm]{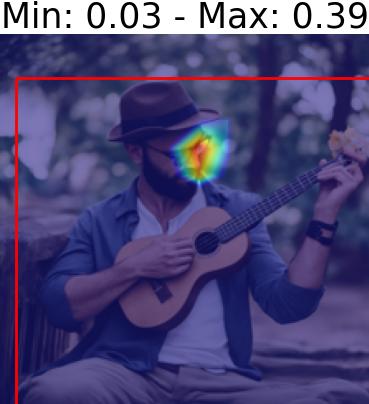}&
\hspace{-0.45cm} \includegraphics[align=c, width=1.54cm]{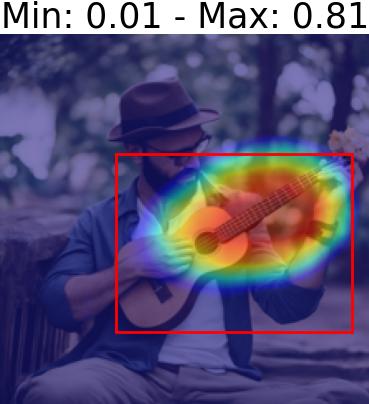}&
\hspace{-0.45cm} \includegraphics[align=c, width=1.54cm]{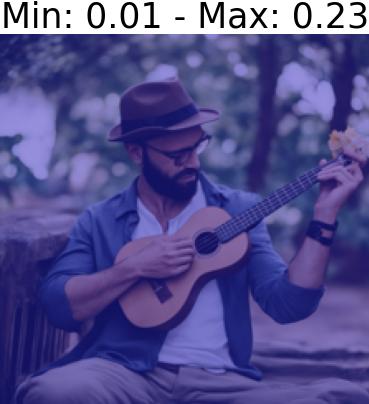}&
\hspace{-0.45cm} \includegraphics[align=c, width=1.54cm]{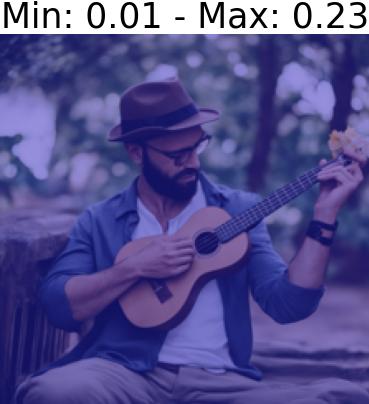}&
\hspace{-0.45cm} \includegraphics[align=c, width=1.54cm]{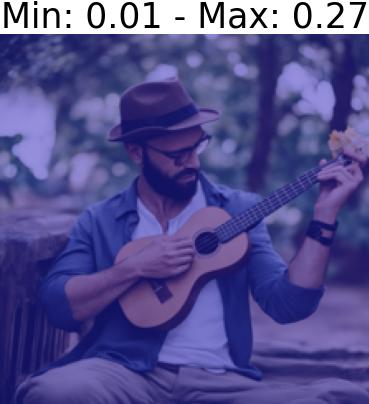}\\
\vspace{0.05cm}
\rotatebox[origin=c]{90}{\scriptsize SSL-Align } & 
\hspace{-0.45cm} \includegraphics[align=c, width=1.54cm]{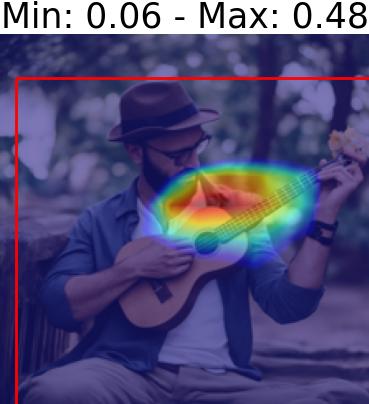}&
\hspace{-0.45cm} \includegraphics[align=c, width=1.54cm]{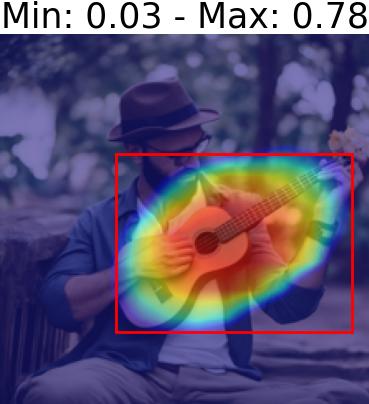}&
\hspace{-0.45cm} \includegraphics[align=c, width=1.54cm]{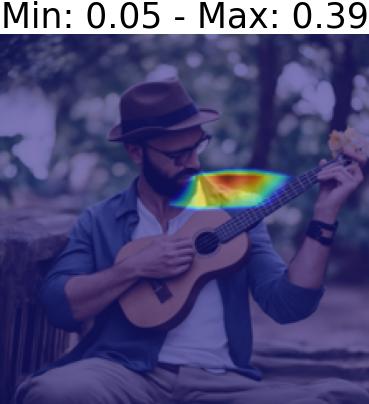}&
\hspace{-0.45cm} \includegraphics[align=c, width=1.54cm]{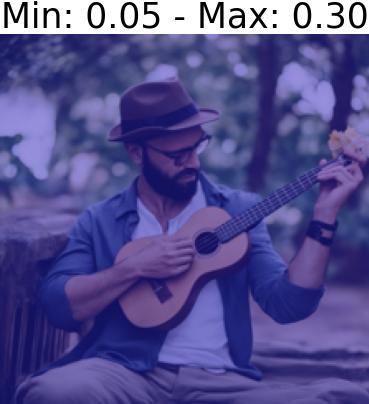}&
\hspace{-0.45cm} \includegraphics[align=c, width=1.54cm]{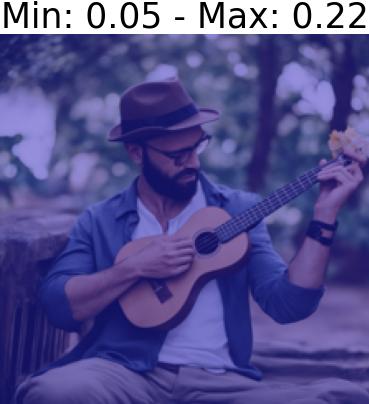}\\
\end{tabular}
\caption{Example of localization results  of the different models in both \textit{positive} and \textit{negative} audio samples in the \new{extended} IS3. The audio-visual similarities below the universal threshold of each model are clipped, then normalized to the interval $[0, 1]$ and overlaid with the original image. The min. and max. values of the audio-visual similarities are reported on top of each image. }
\label{fig:inferences}
\end{figure}

We conducted our experiments on the extended test sets of VGG-SS \cite{LVS} and IS3 \cite{senocak2024aligning} datasets (for VGG-SS we use 3,983 out of 5,000 videos that were available on YouTube). 
Table \ref{tab:all_metrics} contains the results on the \textit{positive} and each of the three \textit{negative} audio inputs as well as the global metrics for both types of inputs, the best results are in bold and the second best underlined.
\new{All metrics except for two have been computed with our proposed universal threshold, U. th.,  (introduced in Section \ref{sec:method}). On the other hand, cIoU Adap. and AUC Adap. use the  adaptive threshold proposed in \cite{senocak2024aligning}}.
 In the case of \textit{positive} audio input and IS3 test set, the first and second best models are SSL-Align and SSL-TIE, respectively, regardless  of the  threshold used to get the localization maps. 
 In contrast, in VGG-SS, the best model with the universal threshold is SSL-TIE and SLAVC second. With the adaptive threshold, \ie the one that uses the prior knowledge of the sounding object's size, the best results are achieved by SSL-Align for cIoU  and by SSL-TIE for AUC. 
EZ-VSL is the best in \textit{silence} in VGG-SS and second best in IS3, while SSL-TIE is best in IS3 and second best in VGG-SS. LVS is best in \textit{noise} and \textit{offscreen} for both datasets. SSL-Align is second best in \textit{noise} in VGG-SS and SSL-TIE in IS3. EZ-VSL is second best in \textit{offscreen} in IS3 and SSL-Align in VGG-SS. Although, LVS and EZ-VSL have poor results in \textit{positive} audio inputs, thus providing the worst global metrics. 
The global performance for both \textit{positive} and \textit{negative} audio is measured by F$_\textbf{LOC}$ and F$_\textbf{AUC}$. We observe that best overall model in VGG-SS is SSL-TIE and in IS3 is SSL-Align, while SSL-TIE is the second best in IS3 and SLAVC in VGG-SS. 
The results in Table \ref{tab:all_metrics} are consistent with the analysis shown in Figure \ref{fig:boxplot}, where the best models according to the global metrics are the ones with less overlap in the IQR of maximum similarity values in the \textit{positive} and \textit{negative} cases\new{, \ie the more discriminatory ones.} 

A qualitative example is shown in Figure \ref{fig:inferences}. Each row represents a different model, and each column corresponds to different audio inputs paired with the same image. The example is from IS3 and there are two different positive audio files: `man', `ukulele' and the three types of \textit{negative} audio samples. 
To show the effect of the universal threshold we set to zero the audio-visual similarity values below that threshold. The resulting values are then normalized to the range $[0, 1]$ and they are combined with the original image. 
Although SSL-Align is the best overall model in IS3, it fails in this example with the audio corresponding to `man' and `silence'. SSL-TIE correctly localizes part of the regions corresponding to both \textit{positive} audio and produces empty maps for the \textit{negative} audio samples. EZ-VSL, SLAVC and FNAC produce empty maps for the \textit{negatives} as well as for the man. In contrast, LVS highlights almost the same region for three different audio inputs: `man', `ukulele' and `silence'. Finally, RCGrad outputs the same localization regardless of the audio input.

\begin{figure}[htb]
   \centering
\includegraphics[width=\columnwidth]{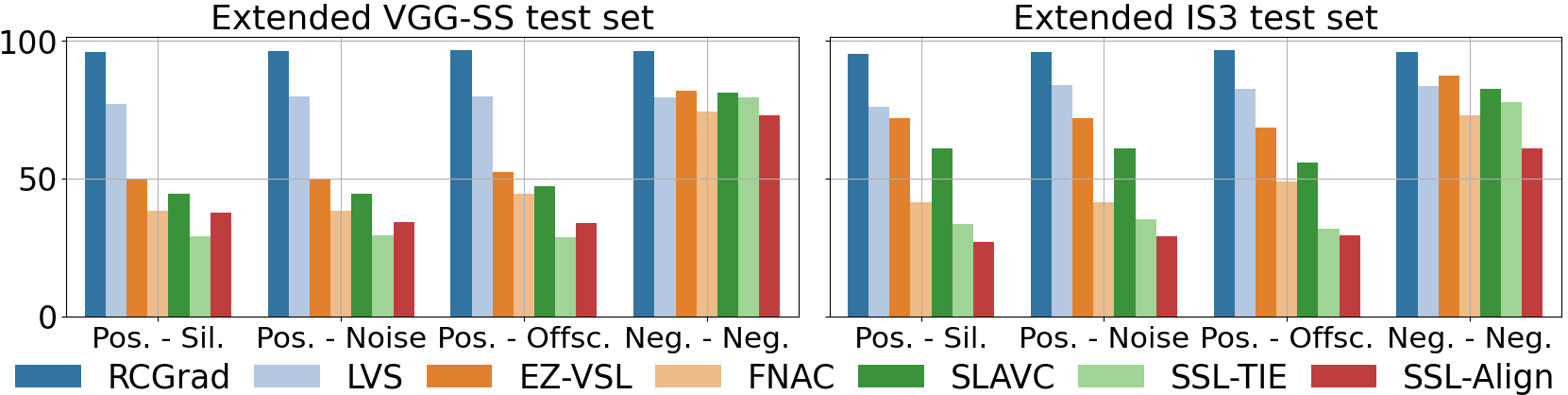}
\caption{IoU between different pairs of localization maps: \textit{positive} and each type of \textit{negatives} 
(\textit{silence}, \textit{noise} and \textit{offscreen} sound) and between the \textit{negatives} themselves. Results of different models depicted in different colors.}
\label{fig:cross_iou}
\end{figure}

Figure \ref{fig:cross_iou} shows the IoU values computed between different pairs of localization maps  (\textit{positive} vs. each type of \textit{negatives} and \textit{negatives} vs. \textit{negatives}). In an effective model, IoU values should be low for \textit{positive} vs. \textit{negative} and high for \textit{negative} vs. \textit{negative}, since it should only identify sounding regions in \textit{positive} cases and produce empty maps for \textit{negative} ones. In VGG-SS, SSL-TIE presents the lowest IoU for \textit{positive} vs. \textit{negatives} (and a more pronounced difference with respect to the IoU of \textit{negatives} vs. \textit{negatives}). In IS3, according to these same observations,  SSL-Align is the best and SSL-TIE is the second best. These results are in accordance with the F values reported in Table \ref{tab:all_metrics}.

\section{Conclusion}
\label{sec:conclusion}

In this paper we assessed the performance of various SOTA Visual Sound Source Localization models, not only in the \textit{positive} audio case, which is the common practice, but also complementing the evaluation on three types of negative audio inputs: \textit{silence},  \textit{noise}  and  \textit{offscreen} sound.  
We proposed new metrics to assess the localization for \textit{negative} audio samples and two new  metrics (F$_{\text{LOC}}$ and F$_{\text{AUC}}$) which evaluate the  performance globally in both \textit{positive} and \textit{negative} cases.  
Additionally, we have established a universal threshold for performing sound localization in an in-the-wild scenario, where no prior knowledge about the sounding object visibility and size is assumed.
Through qualitative and quantitative testing across the VGG-SS and IS3 extended test sets, we identified the models that  better localize the regions where the sound comes from, while being robust to negative audio inputs, although no model was best in all cases. 

Our results highlight the need for further improvements in VSSL models, especially for them to work in real-world scenarios where sound sources may be non present in the image. 
Future work should extend these evaluations to more complex scenes with multiple simultaneous sound sources and consider the temporal dimension of videos.

\end{document}